\documentclass[sigconf]{acmart}
\AtBeginDocument{%
  }

\usepackage{booktabs}
\usepackage[svgnames]{xcolor}
\usepackage{amsmath}
\usepackage{siunitx}
\usepackage{multirow}
\usepackage{hypernat}
\usepackage{enumitem}
\usepackage{amsmath}
\setcopyright{acmlicensed}
\copyrightyear{2025}
\acmYear{2025}
\acmDOI{XXXXXXX.XXXXXXX}
\acmConference[ICAIF '25]{ACM AI in Finance}{Nov 15--18,
  2025}{Singapore}
\acmISBN{978-1-4503-XXXX-X/2018/06}

\begin{document}


\title{Curriculum-Guided Reinforcement Learning for Synthesizing Gas-Efficient Financial Derivatives Contracts}

\author{Maruf Ahmed Mridul}
\affiliation{%
    \institution{Rensselaer Polytechnic Institute}
    \city{Troy}
    \state{New York}
    \country{USA}
    }
\email{mridum@rpi.edu}

\author{Oshani Seneviratne}
\affiliation{%
	\institution{Rensselaer Polytechnic Institute}
	\city{Troy}
	\state{New York}
	\country{USA}
}
\email{senevo@rpi.edu}
\orcid{0000-0001-8518-917X}

\renewcommand{\shortauthors}{Mridul and Seneviratne}

\begin{abstract}
Smart contract-based automation of financial derivatives offers substantial efficiency gains, but its real-world adoption is constrained by the complexity of translating financial specifications into gas-efficient executable code.
In particular, generating code that is both functionally correct and economically viable from high-level specifications, such as the Common Domain Model (CDM), remains a significant challenge. This paper introduces a Reinforcement Learning (RL) framework to generate functional and gas-optimized Solidity smart contracts directly from CDM specifications. We employ a Proximal Policy Optimization (PPO) agent that learns to select optimal code snippets from a pre-defined library. To manage the complex search space, a two-phase curriculum first trains the agent for functional correctness before shifting its focus to gas optimization. Our empirical results show the RL agent learns to generate contracts with significant gas savings, achieving cost reductions of up to 35.59\% on unseen test data compared to unoptimized baselines. This work presents a viable methodology for the automated synthesis of reliable and economically sustainable smart contracts, bridging the gap between high-level financial agreements and efficient on-chain execution.
\end{abstract}




\maketitle

\section{Introduction}

The derivatives market, while a cornerstone of modern finance, is burdened by significant operational complexities, costs, and counterparty risks. 
The lifecycle of these complex agreements, from trade confirmation to settlement and termination, traditionally relies on a web of intermediaries and manual processes, leading to inefficiencies and disputes. Smart contracts ~\cite{szabo1997formalizing}, self-executing agreements deployed on blockchains, offer a transformative solution by enabling the automated, transparent, and immutable execution of financial logic. By codifying a derivative contract on-chain, parties can drastically reduce reliance on intermediaries, minimize settlement risk, and achieve transparency in post-trade processing.

The critical challenge, however, has been bridging the gap between complex financial specifications and executable, on-chain code. The \textbf{CDM}, an industry standard for representing financial products \cite{finos_cdm}, provides the structured, unambiguous, high-level specification that makes this automation feasible. While a direct, rule-based translation from a CDM specification to a functional smart contract is possible, it ignores a crucial factor for real-world adoption: \textit{economic viability}. Every computation on a blockchain incurs a cost, colloquially known as \textbf{gas}. For a smart contract to be practical, its deployment and execution costs must be minimal. Inefficient code leads not only to prohibitive transaction fees for end-users but also impacts the overall throughput and scalability of the network \cite{chen2020gaschecker}. Therefore, the central problem is not merely generating a functional contract, but synthesizing one that is \textit{highly gas-efficient} from the ground up, ensuring its economic sustainability.

This optimization task is not trivial. The search space for a gas-optimal implementation is vast; different choices in data types, function implementations, or control flow logic can lead to significant differences in gas costs. Identifying the optimal combination of these elements is very difficult for deterministic algorithms. This paper introduces a novel RL framework designed to tackle this complex generation and optimization task. We model contract generation as a learning process where a PPO RL agent~\cite{schulman2017proximal} learns to assemble contracts by selecting optimal code snippets from a pre-defined  library. The inherent difficulty lies in the vast and sparse search space: most combinations of snippets result in invalid code, making it difficult for the agent to receive a meaningful reward signal to guide its learning. To overcome this, we employ a two-phase curriculum: the agent first learns to generate functionally correct contracts, and only then does its training objective shift to aggressively minimizing gas costs. This tiered approach allows the agent to effectively discover non-obvious, gas-saving implementation patterns.

The primary contributions of this work include: 1) an end-to-end pipeline that automates the generation of gas-optimized smart contracts directly from high-level CDM specifications; 2) the design of a two-phase learning curriculum that makes the problem tractable by training the agent to first master valid code generation before shifting its focus to gas optimization; and 3) an empirical validation showing that our RL agent generated, optimized contracts are significantly more gas-efficient than the functionally-correct but unoptimized baselines generated by the agent prior to its gas-optimization training.

\section{Related Work}

Smart contract generation and optimization have evolved, with a growing focus on improving gas efficiency. In decentralized finance (DeFi), platforms like Aave and Synthetix have begun experimenting with on-chain derivatives, where transaction costs driven by gas usage directly impact user participation, liquidity provisioning, and scalability ~\cite{shah2023systematic, rahman2022systematization}. This section reviews recent works, covering both traditional techniques and more recent innovations in the field.

\vspace{-0.5em}

\paragraph{\textbf{Smart Contract Generation:}}
Research in automated smart contract generation has primarily followed two paths: synthesis from formal specifications and generation using Large Language Models (LLMs).  The formal specification approach aims to guarantee correctness by design. Early work explored the semi-automated translation of institutional concepts into computational structures~\cite{frantz2016institutions}. More recent tools automatically translate structured inputs, such as the formal language Symboleo~\cite{rasti2024automated}, document templates~\cite{tateishi2019automatic}, or knowledge graphs encoding domain logic~\cite{van2023translating, van2025semantic}, into executable code. Other methods leverage ontologies to produce contracts for specific domains like clinical trials~\cite{choudhury2018auto} or supply chains~\cite{marchesi2022automatic}. While these methods produce verifiable code, they are typically deterministic and do not explore alternative implementations to optimize for non-functional properties like gas costs. 

In contrast, LLMs offer a more flexible paradigm for generating code from unstructured natural language. LLMs have been used to synthesize contracts from legal documents~\cite{kang2024using}, transform complex financial agreements~\cite{mridul2025ai4contracts}, and translate code between different smart contract languages~\cite{karanjai2024solmover}. However, LLMs struggle with correctness and security aspects of the generated code, as vulnerabilities may arise from incorrect specifications, not just faulty code~\cite{sorensen2024correct}. To mitigate this, frameworks have been developed to guide LLM output using formal structures like Finite State Machines (FSMs), improving reliability~\cite{luo2025guiding}. Despite their flexibility, LLMs are not inherently designed to perform fine-grained resource optimization.

\vspace{-0.5em}

\paragraph{\textbf{Post-Hoc Optimization and Analysis:}}
Given that most generation methods prioritize functional correctness, a separate line of research focuses on analyzing and optimizing existing smart contracts, with a strong emphasis on reducing gas consumption. A variety of static and symbolic analysis tools, including \texttt{GasChecker}~\cite{chen2020gaschecker}, \texttt{GasSaver}~\cite{nguyen2022gassaver}, and \texttt{Slither}~\cite{feist2019slither}, automatically detect known gas-inefficient patterns. More advanced techniques infer parametric gas consumption bounds~\cite{albert2020gasol} or use machine learning to detect expensive code patterns~\cite{li2023gas}. 

Specific patterns have also been targeted, including design catalogues~\cite{marchesi2020design} and optimizations for loops~\cite{nelaturu2021smart} and storage~\cite{li2021gas}.
The fundamental limitation of these post-hoc methods is their scope; they refine an existing implementation rather than influencing high-level design choices during generation, where more significant gas savings can be realized.

\vspace{-0.5em}

\paragraph{\textbf{Reinforcement Learning for Program Synthesis:}}
The limitations of both generation-first and optimization-last approaches highlight the need for methods that can integrate these concerns. RL, with its capacity for solving complex, sequential decision-making problems, has shown promise in related software engineering tasks~\cite{wang2024enhancing}. In general program synthesis, RL has been used to guide the search process of deductive reasoners~\cite{chen2020program} and to improve the functional correctness of code generated by LLMs~\cite{le2022coderl}. Within the smart contract domain specifically, RL has been successfully applied to discrete problems such as automated security repair~\cite{guo2024smart} and formal verification~\cite{liu2022learning}. Further work has focused on creating environments for RL-driven compiler optimization~\cite{bendib2024reinforcement} and applying imitation learning for knowledge distribution~\cite{davarakis2023reinforcement}.

Prior work thus reveals a clear trade-off: generation methods solely focus on correctness, while optimization tools can only perform local refinements on a fixed design. Although RL has been applied to isolated problems, its use to unify correctness and optimization during the initial synthesis phase remains underexplored. Our work addresses this gap, using an RL agent to construct contracts while co-optimizing for both correctness and gas efficiency.

\section{Experimental Setup}
 
\begin{figure}[b]
    \centering
    \begin{minipage}{1\linewidth}
        \centering
        \includegraphics[width=\linewidth]{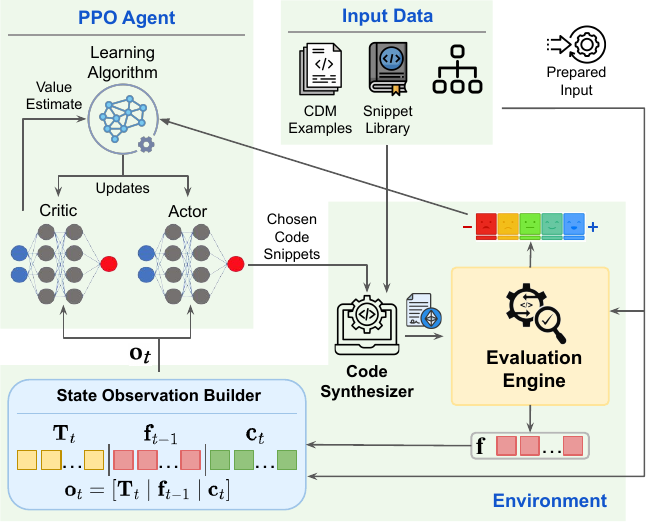}
        (a) Overall Pipeline
    \end{minipage}

    \vspace{1em}

    \begin{minipage}{1\linewidth}
        \centering
        \includegraphics[width=\linewidth]{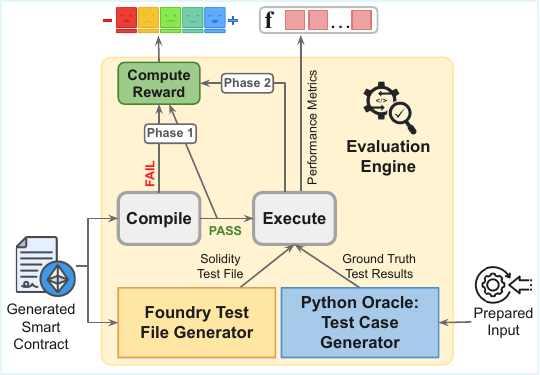}
        (b) Evaluation Engine
    \end{minipage}
    
    \caption{Architecture of the RL framework for generating gas-optimized smart contracts. 
    \normalfont  \\
    (a) The RL agent processes a state observation ($o_t$) containing the contract type ($T_t$), key data from the CDM specification ($c_t$), and performance feedback from the previous generation attempt ($f_{t-1}$). In a single forward pass, the agent's policy selects a complete set of code snippets for all required variables and functions. The Code Synthesizer assembles these snippets into a full smart contract. This contract is then passed to the Evaluation Engine. \\
    (b) The Evaluation Engine uses the Foundry toolchain to verify compilation, functional correctness against oracle-generated tests, and computes gas cost. This analysis produces a reward signal for the agent's policy update, and a performance feedback ($f$) for the next observation.}
    \label{fig:pipeline}
\end{figure}

We implement a comprehensive experimental pipeline to evaluate our approach.
The core of this pipeline involves ingesting high-level financial specifications provided as CDM instances. An RL agent, conditioned on an observation representing the current contract context and its past performance, selects appropriate code snippets from a comprehensive library to programmatically assemble a Solidity smart contract. This generated contract then undergoes a verification process utilizing the Foundry toolchain~\cite{foundry}, encompassing both compilation checks for syntactic and semantic validity and execution of unit tests to ascertain functional correctness. Crucially, gas consumption for both contract deployment and business logic execution is measured during this phase. The results of this verification, including compilation status, test outcomes, and gas costs, are then translated into a reward signal that guides the RL agent's policy refinement in an iterative learning loop. The complete end-to-end architecture is depicted in Figure \ref{fig:pipeline}.

\subsection{RL Input Preparation}
The effectiveness of our RL agent is contingent upon the quality and structure of the data it interacts with. Our system integrates three primary data modalities: CDM instances, reference contract structures, and a curated snippet library.

\vspace{-0.5em}

\subsubsection{Derivative Contract Samples in CDM}
CDM JSON files serve as the primary input. Each file represents a specific instance of a derivative contract, categorized into one of five predefined types: Interest Rate Swap, Equity Swap, Equity Option, Commodity Option, and Foreign Exchange Derivatives. These files contain rich, structured financial and operational metadata (e.g., notional amounts, dates, currencies) essential for generating functionally correct smart contracts. CDM files explicitly do not contain execution rules or logic, acting as a high-level language that defines \textit{what} the financial instrument is, rather than \textit{how} it executes.

Our CDM dataset was constructed ensuring both realism and smart contract compatibility. Initial comprehensive CDM templates were derived using the methodology presented in Mridul et al.'s work on LLM and RAG-powered encoding of financial derivative contracts ~\cite{mridul2025ai4contracts}. Given the complexity and length of these original templates, we then simplified those, retaining only those fields strictly necessary for unambiguous contract execution. Subsequently, while real-life CDM examples are available from sources like the FINOS CDM repository ~\cite{finos-cdm-repository}, their values (e.g., party identifiers, currency specifications) are often not directly compatible with blockchain environments, which require specific formats such as hexadecimal-encoded wallet addresses and on-chain token representations. To address this, we synthetically generated our dataset by populating these simplified CDM templates with relevant values. During this population process, strict adherence to data dependencies and logical consistency was enforced (e.g., ensuring a contract termination date does not precede the original trade date). 
We derived a robust dataset comprising 1,000 samples for each of the five contract types using this process.

\vspace{-0.5em}

\subsubsection{Semantic Mapping}
To translate these CDM contract representations into executable smart contracts, we define a deterministic semantic mapping $
\mu : \mathcal{P} \rightarrow \mathcal{V},
$ where each path \( \mathcal{P} \) in the CDM JSON tree is mapped to a canonical variable \( \mathcal{V} \). For instance:
\[
\mu\left(
\begin{array}{l}
\texttt{trade.tradableProduct....payout.} \\
\texttt{optionPayout[0].buyerSeller.buyer}
\end{array}
\right)
= \texttt{buyer}
\]

Here, the path navigates through the CDM structure to locate the party designated as the \texttt{buyer} in the first option payout leg. The mapping resolves this complex path to a canonical Solidity variable, \texttt{buyer}, which may be used in contract logic to assign rights or enforce conditions (e.g., ensuring only the buyer can trigger a payout or receive funds under specific conditions).

Only variables explicitly listed in the contract type schema are used for a given contract.

\vspace{-0.5em}

\subsubsection{Reference Contract Structures}
These serve as blueprints for each contract type (Interest Rate Swap, Equity Swap, Equity Option, Commodity Option, and Foreign Exchange Derivatives), outlining the essential variables and functions required for a contract of that specific type to be considered complete and functionally correct. Beyond defining structural requirements, these reference structures are instrumental in designing the minimal unit tests used to verify the functional correctness of the contracts generated by the agent.

\vspace{-0.5em}

\subsubsection{Snippet Library}
A comprehensive library of code snippets provides the atomic building blocks from which smart contracts are assembled. This library is extensive, encompassing multiple implementation options for all relevant variables (e.g., for financial amounts, dates, or party addresses) and functions (e.g., for settlement logic or contract initialization). The library was created by deliberately varying variable types and function implementation styles to offer diverse options for the RL agent.

The snippet library includes systematic groupings of variable types and their Solidity declaration variants. For instance, for numeric values like a contract’s \texttt{strikePrice}, options include types of varying bit-width or initialization methods. One snippet might declare it for constructor initialization, such as \texttt{int64 public strikePrice;}, where the value is passed to the contract upon deployment. In contrast, another option allows for direct initialization by the smart contract with a CDM-derived value, or using a different data type for gas optimization, like \texttt{uint256 public strikePrice = <STRIKE\_PRICE\_VALUE>;}. This provides the agent choices for managing memory and deployment costs based on the specific context and value range.

Similarly, the library emphasizes gas and code diversity for function variants. For instance, snippets for settlement functions vary in their use of conditional logic, nested branches, ternary operators, and strategies for intermediate variable computations versus inline expressions. To illustrate, for a price difference calculation within a function, the agent might choose between an explicit \texttt{if-else} block:
\begin{verbatim}
        uint256 diff;
        if (endPrice > startPrice) {
            diff = endPrice - startPrice;
        } else {
            diff = startPrice - endPrice;
        }
\end{verbatim}
or a more concise ternary operator:
\begin{verbatim}
    uint256 diff = endPrice > startPrice ? 
    endPrice - startPrice : startPrice - endPrice;
\end{verbatim}
This diversity allows the agent to explore different computational paths.

\subsection{Reinforcement Learning Framework}

Our RL framework is instantiated with a PPO agent interacting within a custom Gymnasium environment~\cite{towers2024gymnasium}
, specifically tailored for code synthesis.

\subsubsection{Agent and Environment Formulation}
The custom environment extends the \texttt{gymnasium.Env} interface, providing the necessary interaction layer for the RL agent.

\textbf{Observation Space ($\mathcal{O}$):} The agent's observation at timestep $t$, $\mathbf{o}_t \in \mathcal{O}$, is a concatenated vector designed to provide rich contextual information.
\[ \mathbf{o}_t = [\mathbf{T}_t \mid \mathbf{f}_{t-1} \mid \mathbf{c}_t] \]
where:

    \paragraph{ $\mathbf{T}_t \in \mathbb{R}^5$:} A one-hot vector encoding the current contract type. This allows the agent to condition its policy on the specific financial instrument.
    \paragraph{ $\mathbf{f}_{t-1} \in \mathbb{R}^{2 \times |\mathcal{F}|}$:} A performance feedback vector summarizing the outcomes from the previous episode. For each of the $|\mathcal{F}|$ universal functions, this vector includes two components: a binary flag indicating unit test pass/fail and a normalized gas cost. Specifically, if $g_j$ is the gas consumed by function $j$ and $B_{\text{func}}$ is the function gas budget, its normalized value is $\min(g_j/B_{\text{func}}, 1.0)$, transformed to $1.0 - \min(g_j/B_{\text{func}}, 1.0)$ for reward alignment. This provides historical memory for correctness and efficiency.
    \paragraph{ $\mathbf{c}_t \in \mathbb{R}^{10}$:} A CDM feature vector extracting up to ten key numerical features (e.g., trade date, strike price, notional amounts) from the current CDM instance, where the values are normalized to $[0, 1]$. This informs the agent about the magnitude of data for snippet assignments, critical for avoiding data-type overflows.

\textbf{Action Space ($\mathcal{A}$):} The agent's action space is defined as a \texttt{MultiDiscrete} space. Let $\mathcal{S}$ be the universal set of all symbols (variables and functions) available across all contract types. For each symbol $s_i \in \mathcal{S}$, there exists a finite set of available snippet options, $\Omega(s_i)$, with cardinality $|\Omega(s_i)|$. The action space $\mathcal{A}$ is then a vector where each dimension corresponds to the index of a chosen snippet for each $s_i \in \mathcal{S}$:
\[ \mathcal{A} = \text{MultiDiscrete}([\vert\Omega(s_1)\vert, \vert\Omega(s_2)\vert, \ldots, \vert\Omega(s_N)\vert]) \]
where $N = |\mathcal{S}|$. Although the agent outputs a full action vector for all $N$ symbols to maintain a fixed policy dimension, only a contextually relevant subset of actions, corresponding to the required symbols $\mathcal{S}_T$ for the current contract type $T$, is used for contract generation. This design enhances the robustness of the training process and promotes generalization by continuously exposing the agent to the complete range of possibilities.

\subsubsection{PPO Algorithm for Policy Learning}
PPO was chosen for its strong balance of training stability and sample efficiency, which are crucial for our complex generative environment with a sparse reward signal and high-dimensional discrete action space. 
Using PPO, the agent learns an optimal policy $\pi_{\theta}(\mathbf{a}_t|\mathbf{o}_t)$ that maps observations $\mathbf{o}_t$ to a distribution over actions $\mathbf{a}_t$.

At each episode, the PPO agent’s \textit{actor} network, implemented as a Multi-Layer Perceptron (MLP), selects a complete set of code snippets from the library in a single forward pass. These selections are then combined to construct the contract. The \textit{critic} network, also implemented as an MLP, estimates the expected return from $\mathbf{o}_t$. The \texttt{advantage estimate} $\hat{A}_t$ is then computed based on the observed reward and the critic's baseline, indicating how much better the selected action was compared to the baseline. PPO then updates the policy using the clipped surrogate loss ~\cite{schulman2017proximal} to enable stable learning over high-dimensional code snippet selections:
\[ L^{CLIP}(\theta) = \hat{\mathbb{E}}_t \left[ \min\left(r_t(\theta) \hat{A}_t, \text{clip}(r_t(\theta), 1-\epsilon, 1+\epsilon)\hat{A}_t\right) \right] \]

Here, $\theta$ represents the parameters of the policy network, and $r_t(\theta) = \frac{\pi_{\theta}(\mathbf{a}_t|\mathbf{o}_t)}{\pi_{\theta_{old}}(\mathbf{a}_t|\mathbf{o}_t)}$ is the probability ratio between the new and old policies for the selected snippet configuration. The expectation $\hat{\mathbb{E}}_t$ is taken over a batch of full contract generation episodes. The clipping parameter $\epsilon$ prevents excessively large updates to the policy, ensuring training stability.

\subsubsection{Smart Contract Generation}
The generation process begins by combining the selected code snippets with their corresponding CDM-derived values to assemble the Solidity source code. 
For variable declarations, it dynamically determines the Solidity type and variable name from the chosen snippet. If a snippet includes a placeholder (e.g., representing a value to be initialized), the corresponding CDM value is formatted into a Solidity-compatible literal (e.g., \texttt{uint} values are cast to strings, address-like values are formatted as hexadecimal strings or \texttt{bytes20} literals, string values are quoted) and directly embedded. Otherwise, if the variable is to be initialized at deployment, its type and name are added to the contract's constructor parameters, and the corresponding CDM value is passed during instantiation in the test environment. All functions, including the crucial \texttt{balances} mapping, are then assembled into the final contract.

\subsubsection{Smart Contract Verification}
The generated Solidity code is written to a temporary workspace for compilation and testing using the Foundry toolchain~\cite{foundry}.

\begin{itemize}[leftmargin=*]
    \item \textbf{Compilation (\texttt{forge build}):} This command checks for fundamental syntactic and semantic correctness. If \texttt{forge build} fails, the episode is terminated, and a Phase 1 \emph{(negative)} reward is computed based on compiler errors. 

    \item \textbf{Unit Testing and Functional Validation (\texttt{forge test}):} If compilation succeeds, a Solidity test file is generated. This test file dynamically creates an instance of the synthesized contract, populating its constructor arguments with formatted CDM values that align with the Solidity types chosen by the agent. The test suite includes:
    \begin{itemize}[leftmargin=*]
        \item A dedicated test function for accurate measurement of contract deployment gas.
        \item Function-specific tests (e.g., \texttt{test\_settleOption()}): These tests, derived using an internal \texttt{python} based \textbf{oracle} mechanism, predict expected post-execution state changes (e.g., precise balance updates). The generated contract's actual execution is then compared against these oracle-derived outcomes using Solidity assertions.
    \end{itemize}
\end{itemize}
It is important to highlight a crucial aspect of error detection: while initial \texttt{forge build} verifies the code's structure (e.g., syntax, basic type mismatches), functional errors or runtime issues (e.g., an arithmetic overflow if a large numerical CDM value exceeds the capacity of an agent-chosen \texttt{uint64} variable during a calculation, or incorrect logic leading to an assertion failure or \texttt{require} statement revert) are typically identified during the \texttt{forge test} execution. The environment effectively handles both scenarios: a compilation failure from \texttt{forge build} triggers a dedicated compilation error reward path, whereas \texttt{forge test} failures, while indicating functional issues rather than compilation errors, significantly impact the Phase 2 reward signal.

\subsection{Reward Engineering: A Two-Phase Curriculum Learning Strategy}
To address the inherent sparse reward problem in code synthesis, where most randomly generated contracts are invalid, we employ a  \textbf{two-phase curriculum learning strategy} ~\cite{bengio2009curriculum}. 
This two-phase approach effectively mirrors human learning, first mastering fundamental correctness before optimizing for efficiency, enabling effective learning in an otherwise intractable search space.

We decompose the overall objective into two sequential and tractable sub-tasks: first, ensuring syntactic and semantic correctness; and second, minimizing gas consumption through efficiency-driven optimization.

\paragraph{\textbf{Phase 1: Compilation Training}}
The initial phase focuses solely on enabling the agent to generate compilable Solidity code. Reward shaping is utilized to provide a dense signal, even amidst frequent failures.
\begin{itemize}[leftmargin=*]
    \item \textbf{Successful Compilation:} If contract compilation succeeds, the agent receives a constant positive reward, $R_{success}$.
    \item \textbf{Compilation Failure:} If compilation fails, the output from the Solidity compiler is parsed to count the number of distinct error messages, $N_{errors}$. The penalty is formulated as:
    \[ P_{compile\_fail} = P_{base} + (N_{errors} \times P_{per\_error}) \]
    where $P_{base}$ is a base penalty for compilation failure and $P_{per\_error}$ is the designated penalty for each compilation error. This creates a gradient, steering the agent towards reducing errors incrementally. $N_{errors}$ is capped at a minimum of 1 for any failure to ensure a penalty.
\end{itemize}

\begin{table}[b]
  \caption{Parameters of the RL Training Environment}
  \label{tab:parameters}
  \begin{tabular}{llr}
    \toprule
    \textbf{Phase} & \textbf{Parameter} & \textbf{Value} \\
    \midrule
    \multirow{3}{*}{\shortstack[l]{1 : Compilation \\ Training}} & $R_{\text{success}}$ & 200 \\
                       & $P_{\text{base}}$ & -100 \\
                       & $P_{\text{per\_error}}$ & -10 \\
    \midrule
    \multirow{6}{*}{\shortstack[l]{2 : Gas \\ Optimization}} & $R_{\text{max}}$ & 1500 \\
                       & $B_{\text{deploy}}$ & 1,200,000 \\
                       & $B_{\text{func}}$ & 90,000 \\
                       & $w_{\text{deploy}}$ & 0.35 \\
                       & $w_{\text{func}}$ & 0.65 \\
                       & $P_{\text{compile\_fail}}$ & -200 \\
    \bottomrule
  \end{tabular}
\end{table}

\paragraph{\textbf{Phase 2: Gas Optimization}}
Once the agent demonstrates consistent compilability, the environment transitions to this phase, where the reward function's primary objective becomes minimizing gas consumption. A weighted, multi-budget approach is employed to provide a balanced and stable learning signal, preventing a one-time large deployment cost from overshadowing the function execution cost, which is subject to repetition with each invocation. The total reward is scaled by a maximum possible value, $R_{max}$.

The reward functions are designed to transform raw gas costs into a normalized and bounded signal that is ideal for the learning agent. First, the ratio of used gas to its budget ($G/B$) creates a proportional measure of resource consumption. This ratio is then capped at $1.0$, which prevents excessively large penalties for exceeding the budget and destabilizing the learning process. Second, this \textit{budget utilization} score is inverted using ($1 - \text{score}$) so that lower gas usage correctly corresponds to a higher reward. Finally, the result is scaled by the component's weight ($w_{deploy}$ or $w_{func}$)  and the maximum reward ($R_{max}$) to control its relative importance and overall magnitude.

\begin{itemize}[leftmargin=*]
    \item \textbf{Deployment Reward ($R_{deploy}$):} This component is controlled by a weight, $w_{deploy}$, and evaluates the deployment gas ($G_{deploy}$) against a predefined budget, $B_{deploy}$. The reward calculation is:
    $$R_{deploy} = w_{deploy} \cdot R_{max} \cdot \left(1 - \min\left(\frac{G_{deploy}}{B_{deploy}}, 1.0\right)\right)$$
    
    \item \textbf{Function Reward ($R_{func}$):} This component is controlled by a weight, $w_{func}$, and assesses the average gas cost ($G_{func.avg}$) for all logic functions against a budget, $B_{func}$. The weights are set such that $w_{deploy} + w_{func} = 1$, allowing for the prioritization of different aspects of gas efficiency. The reward calculation is:
    $$R_{func} = w_{func} \cdot R_{max} \cdot \left(1 - \min\left(\frac{G_{func.avg}}{B_{func}}, 1.0\right)\right)$$
    This weighting scheme allows for prioritizing function efficiency, acknowledging that cumulative execution costs often exceed one-time deployment costs for long-lived financial instruments.

    \item \textbf{Total Reward:} The final reward for a functionally correct and successfully compiled contract in Phase 2 is the sum of its parts: 
    $R_{total} = R_{deploy} + R_{func}$.

    \item \textbf{Failure Penalty in Phase 2:} 
    As the agent's policy shifts to prioritize gas optimization, it may explore combinations of code snippets that, while potentially efficient, result in unforeseen syntactic errors. If compilation fails during this phase, $P_{compile\_fail}$ is applied with a fixed value. This serves as critical negative feedback to discourage such lapses and maintain compilation correctness as a hard constraint.
\end{itemize}

\section{Results and Discussion}

We analyze the agent's learning progression through its two-phase training curriculum, evaluate the quantitative gas savings achieved across different contract types, and provide a qualitative analysis of the specific code-level optimizations discovered by the agent.

\subsection{Training Performance}

The agent is trained using the two-phase curriculum learning strategy with environment parameters detailed in Table \ref{tab:parameters}. These parameter values were determined through empirical analysis. For example, $P_{per\_error}$ was set to -10, as we observed a maximum of approximately 10 compiler errors in practice. This structure creates a maximum penalty of -200, which directly balances the +200 reward for a successful compilation. Similarly, the gas budgets for deployment and functions were established based on the maximum observed costs from our initial analysis. Alongside these environment settings, the agent's learning process was guided by an entropy coefficient (\texttt{ent\_coef}) that was initially set high to encourage broad exploration of the action space and was gradually reduced in later stages to promote exploitation of high-reward policies.

\begin{figure}[t]
    \centering
    \includegraphics[width=\linewidth]{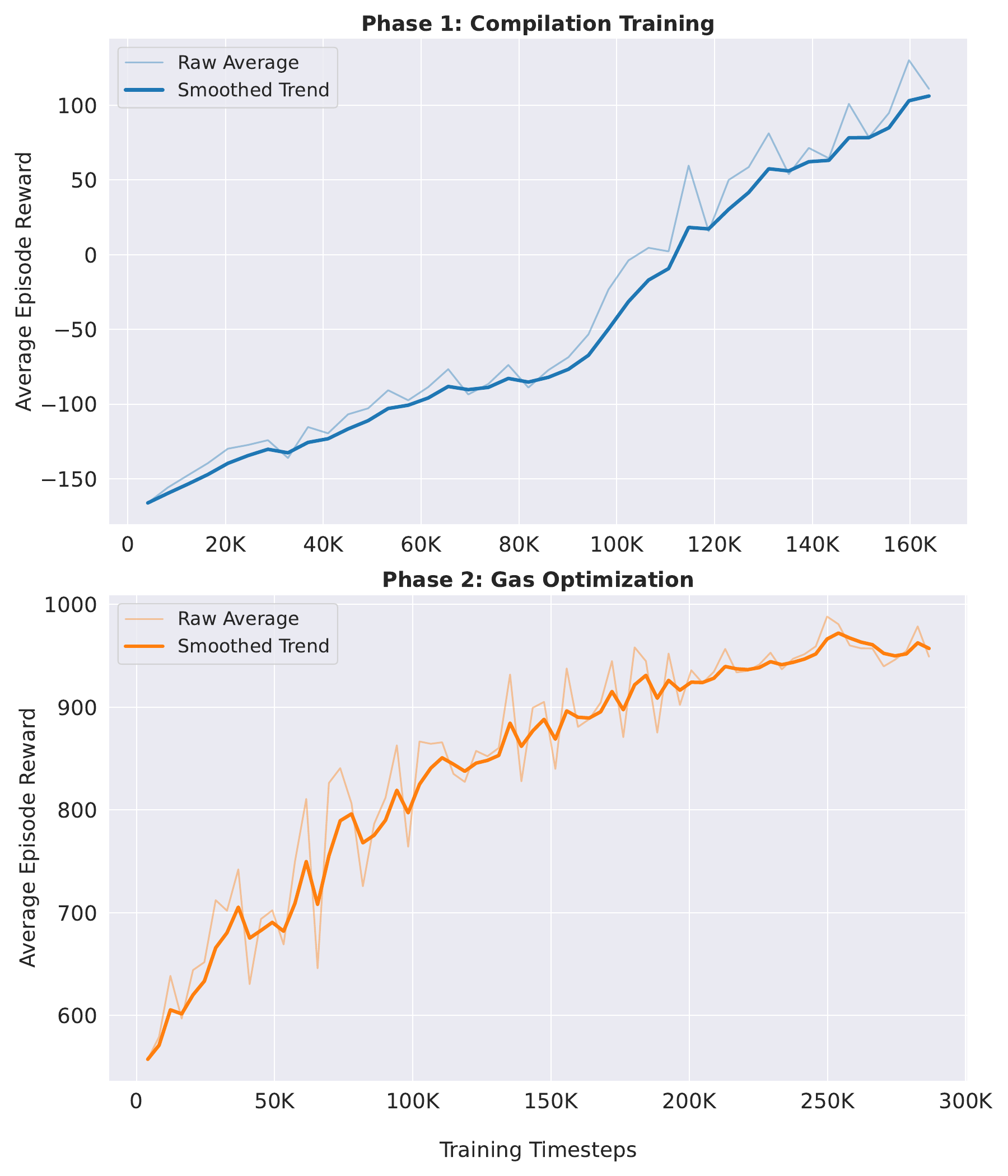}
    \caption{Average episode reward during the two-phase training. 
    \normalfont  The agent first learns to generate compilable code (Phase 1, top) and then learns to optimize for gas efficiency (Phase 2, bottom), as shown by the consistently increasing reward trends.}
    \label{fig:ppo_training}
\end{figure}

The effectiveness of this two-phase curriculum is evident in the agent's learning curves, as shown in Figure \ref{fig:ppo_training}.

    \paragraph{\textbf{Phase 1: Compilation Training}} The top graph shows the agent's progress in learning to generate syntactically correct code. The average episode reward starts at approximately -150, indicating that the agent initially produces code with numerous compilation errors. The smoothed trend line shows a steady, consistent increase, crossing into positive territory around the 110K timestep mark. This demonstrates that the agent is successfully learning to select snippet combinations that result in valid, compilable contracts. Although the reward trend was still rising, Phase 1 was concluded after 160K timesteps because the agent was consistently generating compilable code. Instead of aiming for the reward to plateau, we advanced the training to focus on gas optimization, as our framework is designed to handle the now less-likely event of a compilation failure in Phase 2 as well with a penalty.

    \paragraph{\textbf{Phase 2: Gas Optimization:}} The bottom graph illustrates the agent's performance in the gas optimization phase. The reward immediately shifts to a positive scale, as the agent is already capable of producing valid contracts. The clear upward trend from an initial reward of \textasciitilde600 to a plateau around \textasciitilde950 shows the agent is effectively learning to minimize gas consumption. The learning curve is steep initially, indicating rapid discovery of significant optimizations, and then gradually flattens as the agent converges on a near-optimal policy within the given snippet library. We stopped the training after approximately 290K timesteps as the reward had stabilized, suggesting further improvement was unlikely.

\begin{figure*}[t]
    \centering
    \includegraphics[width=1\linewidth]{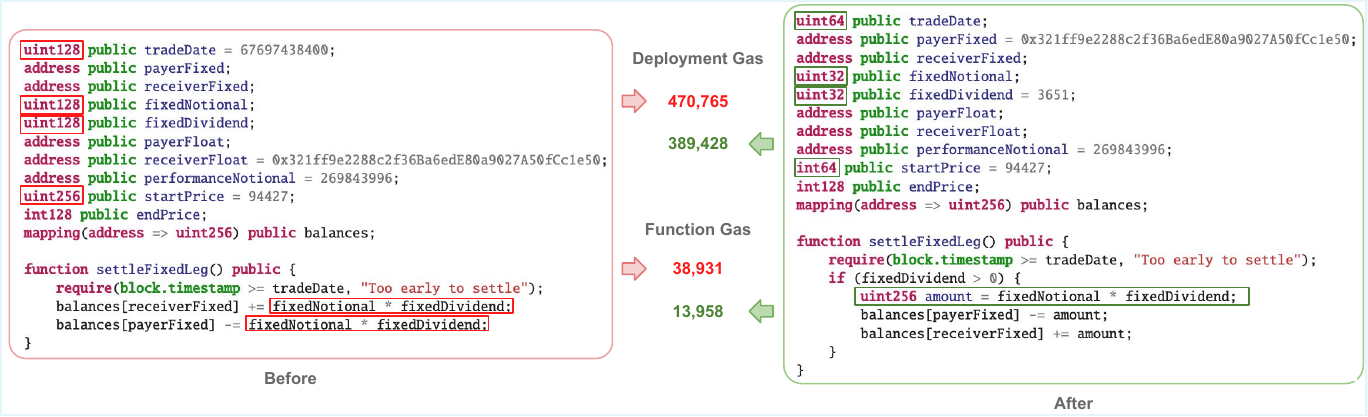}
    \caption{An example of learned optimizations on an Equity Swap contract, with key changes highlighted in \textcolor{red}{red} (before) and \textcolor{DarkGreen}{green} (after) boxes. 
    \normalfont  The agent reduced deployment gas by selecting smaller data types and cut execution gas by eliminating a repeated calculation inside the \textit{settleFixedLeg} function, leading to significant gas savings.}
    \label{fig:contract_comparison}
\end{figure*}

\subsection{Quantitative Gas Optimization Results}

The final performance of the trained agent was evaluated on a test set of 100 previously unseen CDM examples for each of the five types. For this test set, we generated two distinct versions of the smart contracts: a \textit{baseline} version using the agent trained only through Phase 1 (functional but unoptimized), and a final \textit{optimized} version using the fully trained Phase 2 agent. The normalized gas costs of the \textit{optimized} contracts were then compared to those of the \textit{baseline}. The results, summarized in Table \ref{tab:gas_comparison}, demonstrate that the RL agent successfully improved gas efficiency across all categories.

The most significant improvements were observed for the \textit{Equity Swap} and \textit{Interest Rate Swap} contracts, with gas cost reductions of \textbf{21.14\%} and \textbf{35.59\%}, respectively. These contract types are inherently more complex than the others in our dataset, involving more variables and intricate settlement logic. This complexity provided a larger optimization search space and a richer set of diverse code snippets in our library, giving the RL agent more opportunities to discover impactful gas-saving combinations.

In contrast, the \textit{Commodity Option} (1.61\%), \textit{Equity Option} (1.73\%), and \textit{Foreign Exchange} (3.61\%) contracts saw more modest gains. The logic for these contracts is simpler, and the corresponding snippets in our library offered less variation in terms of gas consumption. Nonetheless, the consistent improvement across all types validates our approach. These findings suggest that the degree of optimization achieved by our framework is directly correlated with the richness and diversity of the provided snippet library. A more extensive library with a wider range of gas-differentiated implementation choices would likely yield even greater improvements.

\begin{table}[t]
  \centering
  \caption{
  Comparison of average normalized gas (lower is better) for five smart contract types, showing performance before (\textit{Initial}) and after (\textit{Optimized}) the training process. 
  \normalfont  Averages are computed over 100 \textit{unseen} contracts per type. Normalized gas is a weighted score of deployment gas and average function gas measured against their respective budgets The final column ($\Delta$) quantifies the percentage improvement.
  }
  \label{tab:gas_comparison}
  \begin{tabular}{
    l
    S[table-format=1.4]
    S[table-format=1.4]
    r 
  }
    \toprule
    \multirow{2}{*}{\textbf{Contract Type}} & 
    \multicolumn{2}{c}{\textbf{Avg. Normalized Gas}} & 
    \multirow{2}{*}{\textbf{$\Delta$ (\%)}} \\
    
    \cmidrule(lr){2-3}
    & {\textbf{Initial}} & {\textbf{Optimized}} & \\ 
    \midrule

    CommodityOption & 0.3929 & 0.3865 & \textbf{1.61}~\textcolor{ForestGreen!15}{$\blacktriangle$} \\
    EquityOption    & 0.4022 & 0.3953 & \textbf{1.73}~\textcolor{ForestGreen!25}{$\blacktriangle$} \\
    EquitySwap      & 0.3869 & 0.3051 & \textbf{21.14}~\textcolor{ForestGreen!85}{$\blacktriangle$} \\
    ForeignExchange & 0.2108 & 0.2032 & \textbf{3.61}~\textcolor{ForestGreen!40}{$\blacktriangle$} \\
    InterestRateSwap& 0.4900 & 0.3156 & \textbf{35.59}~\textcolor{ForestGreen!115}{$\blacktriangle$} \\
    \bottomrule
  \end{tabular}
\end{table}

\subsection{Qualitative Analysis of Learned Optimizations}

Beyond quantitative improvements, a qualitative analysis of the generated code provides insight into the specific optimization strategies the agent learned. Figure \ref{fig:contract_comparison} shows a direct comparison of code excerpts from an Equity Swap contract generated before and after the optimization training. The agent learned to reduce gas costs through key strategies, including:

\paragraph{\textbf{Data Type Selection:}} The optimized contract utilizes smaller, more gas-efficient data types. For instance, \texttt{tradeDate} was changed from \texttt{uint128} to \texttt{uint64}, which is sufficient for storing a Unix timestamp. Some other variables - \texttt{fixedNotional}, \texttt{fixedDividend}, and \texttt{startPrice} were downsized as well. Using smaller data types reduces the amount of storage required on the blockchain, leading to a direct reduction in \textbf{deployment gas} from 470,765 to 389,428.

\paragraph{\textbf{Efficient Computation:}} The agent optimized the logic within the \texttt{settleFixedLeg} function. In the initial version, the calculation \texttt{fixedNotional * fixedDividend} was performed twice, resulting in redundant storage reads and computations. The optimized version calculates the result once, stores it in a memory variable (\texttt{amount}), and reuses it. This simple change drastically reduced the function's execution gas from 38,931 to 13,958, as it minimizes expensive storage access operations.

These examples show that the agent is not merely making random changes but is learning context-aware, human-like optimization patterns that lead to verifiably more efficient smart contracts.

\subsection{Scalability Considerations}

A critical component of our framework is the predefined code snippet library, which constitutes the action space for the RL agent. While effective for validating our approach, the manual curation of this library is not scalable for all potential financial instruments. Accordingly, this paper's core contribution is to demonstrate that \textit{given} a sufficiently diverse action space, our RL methodology can effectively navigate it to find functionally correct and gas-optimal solutions.

However, the advent of LLMs presents a compelling path forward: a more advanced, hybrid pipeline where LLMs are used to dynamically generate the action space itself. For any required component, an LLM could be prompted to produce multiple, functionally equivalent implementation variants with diverse coding styles, data types, and logic. Such LLM-generated set of snippets would then serve as the action space for our agent.

Beyond automating the action space, future work could expand the optimization objectives. The agent’s reward function could be augmented to co-optimize for security by integrating static analysis tools to penalize known vulnerabilities, thereby training the agent to generate secure-by-design contracts.

\section{Conclusion}
We demonstrate the use of RL to generate gas-optimized smart contracts from CDM specifications. By training an RL agent with a two-phase curriculum, we ensure both functional correctness and significant gas savings, achieving reductions of up to 35.59\% for complex contract types like \textit{Interest Rate Swaps}. The agent effectively learns strategies for gas optimization, such as selecting more efficient data types and refining computation paths.
These results establish a practical and scalable methodology for translating high-level financial agreements into economically viable on-chain implementations.

\noindent \textbf{Code and Reproducibility:} 
To support reproducibility and future research, we provide our full codebase, evaluation scripts, and our synthetic CDM dataset at this anonymous GitHub repository \url{https://anonymous.4open.science/r/gas-efficient-smart-contract-generation}.

\section*{Acknowledgment}
Authors acknowledge the support from NSF IUCRC CRAFT center research grant (CRAFT Grant \#22022) for this research. The opinions expressed in this publication do not necessarily represent the views of NSF IUCRC CRAFT.

\balance
\bibliographystyle{ACM-Reference-Format}
\bibliography{references}

\end{document}